\documentclass[letterpaper, 10 pt, conference]{IEEEtran}

\IEEEoverridecommandlockouts 

\usepackage{cite}
\usepackage{amsmath,amssymb,amsfonts}
\usepackage{hyperref}
\usepackage[ruled,noend,linesnumbered]{algorithm2e}
\usepackage{listings}
\usepackage{graphicx}
\usepackage{textcomp}
\usepackage{slashbox}
\usepackage{xcolor}
\usepackage{bm}

\title{\LARGE \bf Anomaly Detection with Selective Dictionary Learning}

\author{
    \IEEEauthorblockN{Denis C. ILIE-ABLACHIM} \\
    \IEEEauthorblockA{
        \textit{Faculty of Automatic Control and Computers}\\
        \textit{University Politehnica of Bucharest}\\
        denis.ilie\_ablachim@upb.ro
    }
    \and
    \IEEEauthorblockN{Bogdan DUMITRESCU}\\
    \IEEEauthorblockA{
        \textit{Faculty of Automatic Control and Computers}\\
        \textit{University Politehnica of Bucharest}\\
        bogdan.dumitrescu@upb.ro
    }
    \thanks{
        This work was supported by a grant of the Ministry of Research, Innovation and Digitization, CNCS - UEFISCDI, project number PN-III-P4-PCE-2021-0154, within PNCDI III.
        It  has also been funded by the Ministry of Investments and European Projects through the Human Capital Sectoral Operational Program 2014-2020, Contract no. 62461/03.06.2022, SMIS code 153735.
    }
}

\begin{document}
\maketitle

\begin{abstract}
In this paper we present new methods of anomaly detection based on Dictionary Learning (DL) and Kernel Dictionary Learning (KDL). The main contribution consists in the adaption of known DL and KDL algorithms in the form of unsupervised methods, used for outlier detection. We propose a reduced kernel version (RKDL), which is useful for problems with large data sets, due to the large kernel matrix. We also improve the DL and RKDL methods by the use of a random selection of signals, which aims to eliminate the outliers from the training procedure. All our algorithms are introduced in an anomaly detection toolbox and are compared to standard benchmark results.
\end{abstract}

\section{Introduction}
Dictionary Learning (DL) is a representation learning method which aims to find a sparse representation for a set of signals $\bm{Y}$, represented as a matrix with $N$ columns (signals) of size $m$. The representation is achieved by computing a dictionary $\bm{D}$ of size $m \times n$ and a sparse representation $\bm{X}$ of size $n \times N$ such that a good approximation $\bm{Y} \approx \bm{DX}$ is obtained. Most applications with dictionary learning are in problems with image denoising, inpainting, signal reconstruction, clustering or classification.

In this paper we present novel methods for unsupervised learning, in particular outlier detection, using DL. The main idea is based on finding a suited dictionary, which is capable of well representing most signals in a dataset, while the outlier signals representation should obtain large errors. Considering the number of outliers significantly lower than the rest of the signals, we expect the dictionary optimization to generally follow the directions of the normal signals.
Our developments cover both the standard and the nonlinear (kernel) DL.

Anomaly detection (outlier detection) is the identification of a subset of signals that have a different representation in relation to the rest of the data. There are several successful anomaly detection methods, such as Isolation Forest (IForest) \cite{liu2008isolation}, Minimum Covariance Determinant (MCD) \cite{hardin2004outlier, rousseeuw1999fast}, One-class SVM detector (OCSVM) or Principal Component Analysis (PCA) Outlier Detector \cite{scholkopf2001estimating}.

There are also several successful sparse coding algorithms used for anomaly detection. An idea was presented in \cite{nguyen2011robust, shekhar2012joint}. These methods consider the data representation as a joint sparse linear combination of training data. By following this technique, the authors try to achieve a direct correlation between all the available signals. Naturally, non-correlated signals are considered as being anomalies.
Another example is given in \cite{adler2015sparse}, where the anomalies are identified in terms of deviation from a trained model. This method tries to achieve good sparse representation for jointly distributed signals, while the other independent signals should be isolated.
An overview of DL can be found in \cite{dlaa}.

The paper is organized as follows.
Section \ref{sec:ADDL} introduces a natural way of solving outlier detection problems using DL algorithms.
Section \ref{sec:SDL} formulates a new DL algorithm, called Selective Dictionary Learning, which aims to improve the anomaly detection algorithm by randomly selecting signals for the training procedure in order to discourage dictionary adaptation to outliers.
In Section \ref{sec:RKDL} we present a reduced kernel version of the DL problem and its Selective form.
Section \ref{sec:Exp} contains the experimental results, obtained by running tests on multivariate data and comparing the results with those of methods available in a Python toolkit for outlier detection.

\section{Anomaly Detection via Dictionary Learning}
\label{sec:ADDL}

The DL problem is formulated as following
\begin{equation}
\begin{array}{ll}
\displaystyle\min_{\bm{D}, \bm{X}} & \|\bm{Y} - \bm{D} \bm{X}\|_{F}^{2} \\
\text { s.t. } & \left\|\bm{x}_{\ell}\right\|_{0} \leq s, \ell=1:N \\
& \left\|\bm{d}_{j}\right\|=1, j=1:n,
\end{array}
\label{eq:DL}
\end{equation}
where $\left\|\cdot\right\|_{0}$ represents the $0$-pseudo-norm and $s$ is the sparsity level.

The standard dictionary learning problem can be solved by using simple strategies. In order to overcome the nonconvexity and the huge dimension of the problem, the optimization procedure is organized in two steps. This method is also known as DL by Alternate Optimization. In this way, the problem is divided in two subproblems: sparse coding and dictionary update. By alternating these two stages for a given number of iterations, the method can obtain good local solutions. An iteration consists of computing the sparse representation $\bm{X}$, while the dictionary $\bm{D}$ is fixed, and then successively updating the dictionary columns, named atoms, while the sparse representation is fixed.
For sparse coding we use Orthogonal Matching Pursuit (OMP) \cite{omp}.
For the dictionary update we use the Approximate version of the K-SVD algorithm (AK-SVD) \cite{k-svd, rubinstein2008efficient}, which optimizes the atoms and their representations successively.

A simple strategy for anomaly detection is to compute the representation error
\begin{equation}
\bm{E} = \bm{Y} - \bm{D} \bm{X}
\end{equation}
and identify the signals that obtain bad representations.
The score of signal $i$ is simply the norm $\|\bm{e}_i \|$
of the $i$-th column of $\bm{E}$.
The largest the error norm, the more likely that signal is an outlier.
The underlying assumption is that signals that are alike can be better represented
by the dictionary designed when solving \eqref{eq:DL}.
However, the dictionary size $n$ and the sparsity level $s$ must be
taken smaller than usual, otherwise the representation may be uniformly good
for all signals and even outliers can be well represented.
A small dictionary favors good representations for signals that are similar,
tuning the atoms for this purpose; a bad representation of the outliers
has little effect on the objective of \eqref{eq:DL}, since they are few.
This trade-off is naturally obtained during the optimization.

Of course, since sparse representation is linear, similarity and dissimilarity
can be thought in terms of direction.
Normal signals belong to a small number of low dimensional subspaces
and the outliers lie on very different subspaces.
This is a model that is appropriate for some anomaly detection problems
but not suited for others.

\section{Selective Dictionary Learning}
\label{sec:SDL}

In the standard DL algorithm, during the training procedure, both stages could be affected by the presence of outliers in the training dataset. The problem of anomaly detection can be solved more easily if we could train the dictionary only on normal data. By neglecting the outliers from the training dataset, we expect to obtain higher representation errors for anomalies.
However, this is not possible, since we do not know which signals are normal
and which are outliers.

To describe our strategy for eliminating most of the outliers from the
training process, we introduce two new parameters in the DL algorithm:
$train\_perc$ ({\it training percent}) and
$train\_drop\_prec$ ({\it training dropout percent}).
The first one represents the percent of data that are used during
the sparse coding stage.
At each iteration, we first apply a random sampling on the training data and only
$train\_perc$\% of the signals are used for sparse coding.
In the dictionary update stage, we further drop off $train\_drop\_perc$\% of the signals,
namely those having the worst representations (largest representation errors).
Although the first random selection can eliminate both normal signals and outliers
from a training iteration, the representation of normal signals is less likely
to suffer, since there are still signals in the current training set that
are similar to them.
On the contrary, outliers are more likely to lack good proxies
and so their representation will worsen.
The second random selection, that of signals with bad representation,
aims to directly remove outliers from the training process.
The dictionary will be updated to better represent the signals that
already have good representations.
Hence, again, the outliers representation will worsen, but the representation
of normal signals not present in the current selection will not be
significantly altered.

The DL problem can be formulated by the use of a zero extended permutation
matrix $\bm{P}$ that is modified at each stage and has the role of randomly
selecting the signals:
\begin{equation}
\begin{array}{ll}
\displaystyle \min_{\bm{D}, \bm{X}} & \|\bm{Y} \bm{P} - \bm{D} \bm{X}\|_{F}^{2} \\
\text { s.t. } & \left\|\bm{x}_{\ell}\right\|_{0} \leq s, \ell=1:N \\
& \left\| \bm{d}_{j} \right\|=1, j=1:n.
\end{array}
\label{eq:SDL}
\end{equation}

\section{Reduced Kernel Dictionary Learning}
\label{sec:RKDL}

Linear spaces can usually hinder good representations. In order to overcome this problem, the standard DL can easily be extended to a nonlinear space. This method is called Kernel Dictionary Learning (KDL) and was introduced in \cite{TRS14} and \cite{NPNC13}. By this, we reproject each signal $\bm{y}$ to a nonlinear space $\phi(\bm{y})$, where $\phi(\cdot)$ is a nonlinear function. The dictionary $\bm{D}$ is also extended to $\phi(\bm{Y})\bm{A}$, where $\bm{A}$ is a matrix with unknown coefficients, taking the role of dictionary. The KDL problem is formulated as
\begin{equation}
\begin{array}{ll}
\displaystyle \min_{\bm{A}, \bm{X}} & \|\varphi(\bm{Y}) - \varphi(\bm{Y}) \bm{A} \bm{X}\|_{F}^{2} \\
\text { s.t. } & \left\|\bm{x}_{\ell}\right\|_{0} \leq s, \ell=1:N \\
& \left\| \varphi(\bm{Y}) \bm{a}_{j} \right\|=1, j=1:n.
\end{array}
\label{eq:SKDL}
\end{equation}
The KDL problem can be solved similarly to the DL problem \eqref{eq:DL}
if Mercer kernels are used, which allows the substitution of a scalar product
of feature vectors $\varphi(\bm{x})^\top \varphi(\bm{y})$ with
a kernel function $k(\bm{x}, \bm{y})$.
However, the problem becomes difficult when using large datasets, due to
the large kernel matrix $\varphi(\bm{Y})^\top \varphi(\bm{Y})$ that results.
The size of the kernel matrix scales linearly with the volume of the data,
which leads to a large volume of memory.
Thus this strategy might not be tractable for problems with large datasets.

In order to overcome this limitation we extend the dictionary $\bm{D}$ to a smaller nonlinear space by $\varphi(\bm{\bar{Y}})\bm{A}$,
where $\bm{\bar{Y}}$ represents a small batch of signals from the original dataset.
Permuting the signals such that $\bm{Y} = [\bm{\bar{Y}} \ \bm{\tilde{Y}}]$,
we can write
\begin{equation}
\varphi(\bm{\bar{Y}}) = \underbrace{[\varphi(\bm{\bar{Y}}) \quad \varphi(\bm{\tilde{Y}})}_{\bm{\varphi(Y)}}] \underbrace{\left[\begin{array}{l}
\bm{I} \\
\bm{0}
\end{array}\right]}_{\bm{P}}.
\label{eq:KP}
\end{equation}

The KDL problem becomes
\begin{equation}
\begin{array}{ll}
\displaystyle \min_{\bm{A}, \bm{X}} & \|\varphi(\bm{Y}) - \varphi(\bm{\bar{Y}}) \bm{A} \bm{X}\|_{F}^{2} \\
\text { s.t. } & \left\|\bm{x}_{\ell}\right\|_{0} \leq s, \ell=1:N \\
& \left\| \varphi(\bm{\bar{Y}}) \bm{a}_{j} \right\|=1, j=1:n.
\end{array}
\label{eq:RKDL-S}
\end{equation}
From \eqref{eq:KP} and \eqref{eq:RKDL-S} we obtain a new optimization problem
\begin{equation}
\begin{array}{ll}
\displaystyle\min_{\bm{A}, \bm{X}} & \|\varphi(\bm{Y}) (\bm{I} - \bm{P A X})\|_{F}^{2}.
\end{array}
\end{equation}
We denote
\begin{equation}
\bm{E} = \bm{I} - \bm{P A X}
\end{equation}
the representation error and
\begin{equation}
\bm{F} = \left[\bm{I} - \bm{P} \sum_{i \neq j} \bm{a}_{i} \bm{x}_{i}^{T}\right]_{I_{j}}
\end{equation}
the representation error without the contribution of the current atom $\bm{a}_{j}$;
by $I_{j}$ we denote the set of signal indices to whose representation
$\bm{a}_{j}$ contributes.
In order to solve the optimization problem \eqref{eq:RKDL-S}, we update the current atom while the other atoms and the representation are fixed.
Removing the index $j$ for a lighter notation, the atom update problem becomes
\begin{equation}
\begin{array}{ll}
\displaystyle \min_{\bm{a}} & \left\|\varphi(\bm{Y})\left(\bm{F} - \bm{P a x}^{\top}\right)\right\|_{F}^{2}.
\end{array}
\end{equation}
Using the trace form of the squared Frobenius norm,
the objective function becomes
\begin{equation}
\begin{array}{l}
\operatorname{Tr}\left[\left(\bm{F}^{\top} - \bm{x a}^{\top} \bm{P}^{\top}\right) \varphi^{\top}(\bm{Y}) \varphi(\bm{Y})\left(\bm{F} - \bm{P a x}^{\top}\right)\right]= \\
=\operatorname{Tr}\left[\bm{F}^{\top} \bm{K F}\right] - 2 \bm{x}^{\top} \bm{F}^{\top} \bm{K P a} + \|\bm{x}\|^{2} \bm{a}^{\top} \bm{P}^{\top} \bm{K P a}.
\end{array}
\label{eq:trK}
\end{equation}
We compute the partial derivative of the objective function with respect to the current atom
\begin{equation}
\frac{\partial(\cdot)}{\partial \bm{a}}=2\|\bm{x}\|^{2} \underbrace{\bm{P}^{\top} \bm{K P}}_{\bar{\bm{K}}} \bm{a} - 2\underbrace{\bm{P}^{\top} \bm{K}}_{\hat{\bm{K}}^{\top}} \bm{F x}
\end{equation}
and so the optimal atom is
\begin{equation}
\bm{a} = \left(\|\bm{x}\|^{2} \bar{\bm{K}}\right)^{-1} \hat{\bm{K}}^{\top} \bm{F x}.
\end{equation}
The atom is normalized after each update; note that the normalizing factor is
$
\left(\bm{a}^{\top} \bar{\bm{K}} \bm{a}\right)^{\frac{1}{2}}
$
in order to obtain $\left\|\bm{a}_{j}\right\|=1$, as required
by the original DL problem.

We call Reduced Kernel Dictionary Learning using a Sampled kernel (RKDL-S)
the method solving problem \eqref{eq:RKDL-S}
and summarize its update step in Algorithm \ref{alg:RKDL-S}.
The optimal representation from step 6 is computed by setting to zero
the partial derivative of \eqref{eq:trK} with respect to $\bm{x}$.
The sparse representation step, not listed here, is made using the
Kernel OMP algorithm \cite{NPNC13}.

\begin{algorithm}[ht!]
\KwData{reduced kernel matrix $\bar{\bm{K}} \in \mathbb{R}^{p \times p}$ \\ \hspace{0.85cm}
partial kernel matrix $\hat{\bm{K}} \in \mathbb{R}^{N \times p}$ \\ \hspace{0.85cm}
current dictionary $\bm{A} \in \mathbb{R}^{N \times n}$ \\ \hspace{0.85cm}
representation matrix $\bm{X} \in \mathbb{R}^{n \times N}$}
\KwResult{updated dictionary $\bm{A}$}
Compute error $\bm{E}=\bm{I}-\bm{P} \bm{A} \bm{X}$ \\
\For{$j=1$ {\bf to} $n$}{
Modify error: $\bm{F}=\bm{E}_{\mathcal{I}_{j}} + \bm{P} \bm{a}_{j} \bm{X}_{j, \mathcal{I}_{j}}$  \\
Update atom: $\bm{a}_{j}= \left(\|\bm{x}\|_{2}^{2} \bar{\bm{K}}\right)^{-1} \hat{\bm{K}}^{\top} \bm{F} \bm{X}_{j, \mathcal{I}_{j}}$ \\
Normalize atom: $\bm{a}_{j} \leftarrow \bm{a}_{j} / \left(\bm{a}_{j}^{\top} \bm{\bar{K}} \bm{a}_{j}\right)^{\frac{1}{2}}$ \\ 
Update representation: $\bm{X}_{j, \mathcal{I}_{j}}^{\top} \leftarrow \bm{F}^{\top} \bm{\hat{K}} \bm{a}_j$ \\
Recompute error: $\bm{E}_{\mathcal{I}_{j}}=\bm{F}-\bm{P}\bm{a}_{j} \bm{X}_{j, \mathcal{I}_{j}}$ \\
}
\caption{RKDL-S}
\label{alg:RKDL-S}
\end{algorithm}

RKDL-S achieves good results, but nevertheless in the training process there are chances to use abnormal signals, by the use of random sampling extraction. This fact can lead to a decrease in accuracy and performance. A better strategy that could overcome this problem would be to use a trained dictionary instead of $\bm{\bar{Y}}$ signals. This can be achieved by using a dictionary, denoted $\bm{\bar{D}}$, obtained from the linear cases in the previous sections. The corresponding optimization problem is
\begin{equation}
\begin{array}{ll}
\displaystyle\min_{\bm{A}, \bm{X}} & \|\varphi(\bm{Y})-\varphi(\bm{\bar{D}})\bm{A} \bm{X}\|_{F}^{2} \\
\text { s.t. } & \left\|\bm{x}_{\ell}\right\|_{0} \leq s, \ell=1:N \\
& \left\|\varphi(\bm{\bar{D}})\bm{a}_{j}\right\|=1, j=1:n.
\end{array}
\label{eq:RKDL-D}
\end{equation}
We name it RKDL-D, the last letter indicating the use of dictionary instead
of sampled signals.
In order to update the current atom, we rewrite the new optimization problem as follows
\begin{equation}
\min _{\bm{a}_{j}}\left\|\varphi(\bm{Y}) - \varphi(\bar{\bm{D}}) \sum_{i \neq j} \bm{a}_{i} \bm{x}_{i}^{\top} - \varphi(\bar{\bm{D}}) \bm{a}_{j} \bm{x}_{j}^{\top}\right\|_{F}^{2}.
\label{eq:atom-RKDL-D}
\end{equation}
Expressing the Frobenius norm via its trace form, \eqref{eq:atom-RKDL-D} becomes
\begin{equation}
\begin{array}{r}
\displaystyle \min_{\bm{a}_{j}} \operatorname{Tr}\left[\left(\varphi^{\top}(\bm{Y}) - \sum_{i \neq j} \bm{x}_{i} \bm{a}_{i}^{\top} \varphi^{\top}(\bar{\bm{D}}) - \bm{x}_{j} \bm{a}_{j}^{\top} \varphi^{\top}(\bar{\bm{D}})\right)\right. \\
\displaystyle \left.\left(\varphi(\bm{Y}) - \varphi(\bar{\bm{D}}) \sum_{i \neq j} \bm{a}_{i} \bm{x}_{i}^{\top} - \varphi(\bar{\bm{D}}) \bm{a}_{j} \bm{x}_{j}^{\top}\right)\right].
\end{array}
\end{equation}
After the substitution of scalar products with the kernel function and negleting the terms that do not depend on $\bm{a}_j$ the final optimization problem is
\begin{equation}
\begin{array}{r}
\displaystyle \min_{\bm{a}_{j}} \operatorname{Tr} \left[ 2 \sum_{i \neq j} \bm{x}_{i} \bm{a}_{i}^{\top} K(\bm{\bar{\bm{D}}}, \bm{\bar{\bm{D}}}) \bm{a}_{j} \bm{x}_{j}^{\top} + \bm{x}_{j} \bm{a}_{j}^{\top} \underbrace{K(\bar{\bm{D}}, \bar{\bm{D}}}_{\bar{K}_{\bar{\bm{D}}}}) \bm{a}_{j} \bm{x}_{j}^{\top} \right. \\ 
\left. - 2 \underbrace{K(\bm{Y}, \bar{\bm{D}})}_{\hat{K}_{\bar{\bm{D}}}} \bm{a}_{j} \bm{x}_{j}^{\top} \right].
\end{array}
\label{eq:trace-RKDL-D}
\end{equation}

Algorithm \ref{alg:RKDL-S} can be easily modified for solving \eqref{eq:trace-RKDL-D},
following the same line of reasoning as above.
In particular, the atom update relation is
\[
\bm{a}_{j}= \left(\|\bm{x}\|_{2}^{2} \bar{\bm{K}}_{\bar{\bm{D}}}\right)^{-1} (\hat{\bm{K}}^{\top}_{\bar{\bm{D}}} + \bar{\bm{K}}_{\bar{\bm{D}}} R) \bm{X}_{j}
\]
and the representation update is
\[
\bm{X}_{j}^{\top} \leftarrow (\bm{\hat{K}}_{\bar{\bm{D}}} - R \bar{\bm{K}}_{\bar{\bm{D}}}) \bm{a}_j,
\]
where we denoted $\bar{\bm{K}}_{D}$ the reduced kernel matrix $k(\bar{\bm{D}}, \bar{\bm{D}})$, $\hat{\bm{K}}_{D}$ the partial kernel matrix $k(\bm{Y}, \bar{\bm{D}})$ and $\bm{R} = \bm{X}^{\top} \bm{A}^{\top} - \bm{X}_{j} \bm{a}_{j}^{\top}$ the transposition representation product with respect to the current atom $\bm{a}_{j}$. The new method is summarized in Algorithm \ref{alg:RKDL-D}.

\begin{algorithm}[ht!]
\KwData{reduced kernel matrix $\bar{\bm{K}}_{\bar{\bm{D}}} \in \mathbb{R}^{p \times p}$ \\ \hspace{0.85cm}
partial kernel matrix $\hat{\bm{K}}_{\bar{\bm{D}}} \in \mathbb{R}^{N \times p}$ \\ \hspace{0.85cm}
current dictionary $\bm{A} \in \mathbb{R}^{N \times n}$ \\ \hspace{0.85cm}
representation matrix $\bm{X} \in \mathbb{R}^{n \times N}$}
\KwResult{updated dictionary $\bm{A}$}
Compute sum $\displaystyle\bm{S}=\sum_{i=1}^{n}\bm{X}_{i}^{\top}\bm{a}_{i}^{\top}$ \\
\For{$j=1$ {\bf to} $n$}{
Modify sum: $\bm{R}=\bm{S} - \bm{X}_{j} \bm{a}_{j}^{\top}$  \\
Update atom: $\bm{a}_{j}= \left(\|\bm{x}\|_{2}^{2} \bar{\bm{K}}_{\bar{\bm{D}}}\right)^{-1} (\hat{\bm{K}}_{\bar{\bm{D}}}^{\top} + \bar{\bm{K}}_{\bar{\bm{D}}}R) \bm{X}_{j}$ \\
Normalize atom: $\bm{a}_{j} \leftarrow \bm{a}_{j} / \left(\bm{a}_{j}^{\top} \bm{\bar{K}}_{\bar{\bm{D}}} \bm{a}_{j}\right)^{\frac{1}{2}}$ \\ 
Update representation: $\bm{X}_{j}^{\top} \leftarrow (\bm{\hat{K}}_{\bar{\bm{D}}} - R \bar{\bm{K}}_{\bar{\bm{D}}}) \bm{a}_j$ \\
Recompute error: $\bm{S} = \bm{R} + \bm{X}_{j} \bm{a}_{j}^{\top}$ \\
}
\caption{RKDL-D}
\label{alg:RKDL-D}
\end{algorithm}

Following the same strategy presented in Section \ref{sec:SDL}, the RKDL methods can easily be adapted to their Selective form. The Selective Reduced Kernel Dictionary Learning (SRKDL) problem is solved as the previous one, by introducing two additional steps for the randomly selection of signals, one for the kernel OMP subproblem and the second one for the matrix coefficients update subproblem. In both cases the random sampling selection is made according to the entire data set (including the abnormal signals).

\section{Experiments}
\label{sec:Exp}

In this section we present the main results obtained with the proposed DL algorithms
for anomaly detection.
All algorithms have been developed in Python and have been introduced in the framework of the PyOD \cite{zhao2019pyod} anomaly detection toolbox. For the evaluation, all vectors of a dataset were normalized and were split into two sets: 60\% for training and 40\% for testing. Each experiment was repeated ten times independently with random splits. In terms of performance, we measure and compute the mean of the area under the receiver operating characteristic (ROC) curve and the precision @ rank n score. We used 16 real-world datasets from different domains, more precisely those gathered in ODDS (Outlier Detection DataSets)\footnote{http://odds.cs.stonybrook.edu/}
and used as benchmark in PyOD, 
and 2 synthetic datasets.

All the algorithms were implemented in Python on a Desktop PC with Ubuntu 20.04 as operating system, having a processor of base frequency of 2.90 GHz (Max Turbo Frequency 4.80 GHz) and 80GB RAM memory (although a 16 GB RAM memory is sufficient). During the experiments, ten different rounds were run. The execution time, receiver operating characteristic value and precision n score were measured based on the average of all rounds. For the nonlinear versions we used two different kernels: radial basis function kernel $k(\bm{x},\bm{y}) = \exp{(-\gamma||\bm{x}-\bm{y}||_2^2)}$ and polynomial kernel $k(\bm{x},\bm{y})=(\gamma  \bm{x}^{\top}\bm{y}+\alpha)^\beta$. The hyperparameters of the kernel functions were chosen according to a grid search. Based on the average results on all the datasets, they were set as following: $\gamma = 1 / m$, $\alpha = 1$ and $\beta = 3$, for the synthetic datasets, while for the rest we used $\gamma = 0.1 / m$ for the rbf kernel and $\gamma = 10 / m$ for the polynomial kernel; we remind that $m$ is the size of a signal. All the implementations are available at \href{https://github.com/denisilie94/pyod-dl}{https://github.com/denisilie94/pyod-dl}, including the two synthetic datasets.

The first synthetic dataset was generated based on two different sparse coded sets of signals. Using two dictionaries, $D_i$, the dictionary for inliers, and $D_o$, the dictionary for outliers, two sets of signals were generated having the sparsity constraint $s=4$. For the numerical experiment we set the number of inliers $N_i = 512$ and number of outliers $N_o = 64$, while the dictionary size are $n_i = 50$ and $n_o = 400$. The signals size was set to $m = 64$. For the outliers signals we used an overcomplete dictionary, since its representation ability is much more diverse than in the case of the dictionary for inliers.
The second dataset consists of random samples from two normal (Gaussian) distributions, of different mean and standard deviation.
We kept the same number of normal and abnormal signals of size 64 as in the previous dataset. The two Gaussian distributions were generated so that the distribution of normal signals clearly overlaps with the distribution of abnormal signals. More exactly the inlier mean and variance are $\mu_{i}=0$ and $\sigma_{i}=0.5$, while the outliers parameters are $\mu_{o}=-0.1$ and $\sigma_{o}=0.45$.

For the DL methods we used small dictionaries of size $n = 50$, while the sparsity constraint was $s=5$. All the dictionaries were trained using $20$ iterations using the AK-SVD method. For the SDL method the $train\_perc = 0.7$ and $train\_drop\_perc = 0.4$. 
For the RKDL method, the size of the matrices $\bm{\bar{Y}}$ from \eqref{eq:RKDL-S} and $\bm{\bar{D}}$ from \eqref{eq:RKDL-D}
was set to $10\%$ of the number of signals. The selective version of RKDL used the parameters $train\_perc = 0.8$ and $train\_drop\_perc = 0.3$.

The results show the good behaviour of our algorithms in detecting outliers via sparse coding. In terms of performance, the DL methods obtain competitive results. The main results are summarized in Tables \ref{tab:roc-pyod}, \ref{tab:prc-pyod} for the public PyOD methods and in Tables \ref{tab:roc-dl}, \ref{tab:prc-dl} for the DL methods. In all the tables we highlight the best three results from both sets of methods (PyOD and DL) taken together. For the synthetic datasets, we noticed that the PyOD methods do not obtain good results. The DL methods obtain better classification results for the dataset generated with sparse coding and the dataset with Gaussian distribution. For ODDS datasets, the overall results are predominantly better for PyOD methods. However, there are a few datasets where DL methods stand out as being better. For example, for the \textit{cardio} dataset, the DL methods achieve the third place in top, while for the \textit{ionosphere} and \textit{satellite} datasets it occupies the second and third place. An interesting dataset is \textit{vertebral} where the DL methods are the best, occupying all three positions of the top. 

\begin{table*}[!htp]
\scriptsize
\centering
\begin{tabular}{|c|c|c|c|c|c|c|c|c|c|c|c|c|c|}
\hline
Data & Samples & Dim. & Out. Perc. & ABOD & CBLOF & FB & HBOS & IForest & KNN & LOF & MCD & OCSVM & PCA \\ \hline
dl\_out & 576 & 64 & 11.1111 & 0.84496 & 0.52271 & 0.58823 & 0.48706 & 0.50225 & 0.56253 & 0.59419 & \textbf{0.8734} & 0.5106 & 0.49162 \\ \hline
2gauss\_out & 576 & 64 & 11.1111 & 0.00633 & 0 & 0.19077 & 0.47946 & 0.28293 & 0 & 0.26359 & 0.00015 & 0.36793 & 0.54262 \\ \hline
arrhythmia & 452 & 274 & 14.6018 & 0.76875 & 0.78382 & 0.77807 & \textbf{0.82193} & \textbf{0.7996} & \textbf{0.7861} & 0.77866 & 0.77897 & 0.78116 & 0.7815 \\ \hline
cardio & 1831 & 21 & 9.6122 & 0.56917 & 0.81003 & 0.58673 & 0.8351 & 0.91844 & 0.72363 & 0.57357 & 0.82715 & \textbf{0.93484} & \textbf{0.95038} \\ \hline
glass & 214 & 9 & 4.2056 & 0.79507 & 0.84125 & \textbf{0.87261} & 0.73887 & 0.74977 & \textbf{0.85076} & \textbf{0.8644} & 0.79006 & 0.63236 & 0.6747 \\ \hline
ionosphere & 351 & 33 & 35.8974 & 0.92476 & 0.89718 & 0.87304 & 0.56144 & 0.85411 & 0.92674 & 0.87535 & \textbf{0.95566} & 0.84192 & 0.7962 \\ \hline
letter & 1600 & 32 & 6.25 & \textbf{0.87825} & 0.78306 & \textbf{0.86605} & 0.59268 & 0.64011 & \textbf{0.87656} & 0.85935 & 0.8074 & 0.61182 & 0.5283 \\ \hline
lympho & 148 & 18 & 4.0541 & 0.91097 & 0.96731 & 0.97528 & \textbf{0.99569} & \textbf{0.99288} & 0.9745 & 0.97709 & 0.91125 & 0.97587 & \textbf{0.98467} \\ \hline
mnist & 7603 & 100 & 9.2069 & 0.78153 & 0.84041 & 0.72046 & 0.57419 & 0.80673 & 0.84813 & 0.71608 & \textbf{0.86661} & \textbf{0.85289} & \textbf{0.85266} \\ \hline
musk & 3062 & 166 & 3.1679 & 0.18444 & \textbf{1} & 0.52626 & \textbf{0.99998} & 0.99984 & 0.79857 & 0.52867 & 0.99997 & \textbf{1} & 0.99995 \\ \hline
optdigits & 5216 & 64 & 2.8758 & 0.46674 & \textbf{0.7692} & 0.44336 & \textbf{0.87325} & \textbf{0.70608} & 0.37076 & 0.45004 & 0.3979 & 0.49972 & 0.50856 \\ \hline
pendigits & 6870 & 16 & 2.2707 & 0.68776 & 0.89307 & 0.45953 & 0.92381 & \textbf{0.94964} & 0.74865 & 0.46975 & 0.83439 & \textbf{0.93031} & \textbf{0.93525} \\ \hline
pima & 768 & 8 & 34.8958 & \textbf{0.67938} & 0.65781 & 0.62345 & \textbf{0.69995} & 0.67798 & \textbf{0.70781} & 0.62705 & 0.67528 & 0.6215 & 0.64811 \\ \hline
satellite & 6435 & 36 & 31.6395 & 0.57137 & 0.74942 & 0.55717 & 0.75811 & 0.6937 & 0.68364 & 0.55727 & \textbf{0.80304} & 0.66224 & 0.59884 \\ \hline
satimage-2 & 5803 & 36 & 1.2235 & 0.81896 & \textbf{0.99922} & 0.45701 & 0.98042 & \textbf{0.99384} & 0.9536 & 0.45774 & \textbf{0.99593} & 0.9978 & 0.98218 \\ \hline
vertebral & 240 & 6 & 12.5 & 0.42615 & 0.43309 & 0.41658 & 0.32625 & 0.39276 & 0.38166 & 0.40811 & 0.39158 & 0.44308 & 0.40269 \\ \hline
vowels & 1456 & 12 & 3.4341 & \textbf{0.96059} & 0.92221 & \textbf{0.94252} & 0.67267 & 0.75966 & \textbf{0.968} & 0.94096 & 0.80761 & 0.78021 & 0.60267 \\ \hline
wbc & 378 & 30 & 5.5556 & 0.90473 & 0.92005 & 0.93254 & \textbf{0.95163} & 0.93073 & \textbf{0.93662} & \textbf{0.93488} & 0.92102 & 0.93189 & 0.91587 \\ \hline
\end{tabular}
\caption{ROC Performance - PyOD methods}
\label{tab:roc-pyod}
\end{table*}

\begin{table*}[!htp]
\scriptsize
\centering
\begin{tabular}{|c|c|c|c|c|c|c|c|c|c|c|}
\hline
Data & DL & SDL & \multicolumn{2}{|c|}{RKDL-S} & \multicolumn{2}{|c|}{RKDL-D} & \multicolumn{2}{|c|}{SRKDL-S} & \multicolumn{2}{|c|}{SRKDL-D} \\ \cline{4-11}
&&& rbf & poly & rbf & poly & rbf & poly & rbf & poly \\ \hline
dl\_out & \textbf{0.89666} & \textbf{0.85336} & 0.36259 & 0.36424 & 0.39653 & 0.37279 & 0.32207 & 0.34155 & 0.3849 & 0.36028 \\ \hline
2gauss\_out & \textbf{0.91278} & \textbf{0.90274} & 0.05688 & 0.02432 & 0.0165 & 0.01193 & \textbf{1} & 0.82143 & 0.00569 & 0.0036 \\ \hline
arrhythmia & 0.77057 & 0.77194 & 0.68557 & 0.63535 & 0.70723 & 0.76356 & 0.72746 & 0.72333 & 0.72112 & 0.78032 \\ \hline
cardio & 0.70023 & 0.72884 & 0.63584 & 0.92797 & 0.60367 & 0.82092 & 0.69322 & \textbf{0.93747} & 0.63524 & 0.89624 \\ \hline
glass & 0.67484 & 0.64841 & 0.77257 & 0.65649 & 0.80463 & 0.68711 & 0.79033 & 0.68143 & 0.62142 & 0.72519 \\ \hline
ionosphere & \textbf{0.93401} & \textbf{0.93923} & 0.87097 & 0.62494 & 0.88993 & 0.80313 & 0.85409 & 0.60376 & 0.8578 & 0.77142 \\ \hline
letter & 0.82366 & 0.81978 & 0.72686 & 0.33016 & 0.74923 & 0.41022 & 0.72863 & 0.3328 & 0.71789 & 0.42632 \\ \hline
lympho & 0.91635 & 0.93615 & 0.54727 & 0.711 & 0.7846 & 0.95522 & 0.60362 & 0.82363 & 0.82788 & 0.9102 \\ \hline
mnist & 0.81029 & 0.79723 & 0.66476 & 0.80017 & 0.54424 & 0.57414 & 0.6443 & 0.71678 & 0.5917 & 0.57642 \\ \hline
musk & 0.88574 & 0.89346 & 0.69204 & 0.75747 & 0.55197 & 0.70492 & 0.77949 & 0.9375 & 0.70581 & 0.82653 \\ \hline
optdigits & 0.40227 & 0.4084 & 0.36615 & 0.41572 & 0.40347 & 0.56437 & 0.39185 & 0.43933 & 0.48264 & 0.61547 \\ \hline
pendigits & 0.62745 & 0.63382 & 0.8356 & 0.91127 & 0.78032 & 0.88282 & 0.83539 & 0.92798 & 0.84152 & 0.91862 \\ \hline
pima & 0.56365 & 0.55959 & 0.60599 & 0.64188 & 0.6131 & 0.65217 & 0.62009 & 0.6357 & 0.63752 & 0.65449 \\ \hline
satellite & 0.65351 & 0.64655 & 0.65338 & 0.66762 & 0.64236 & 0.59866 & \textbf{0.77197} & 0.68252 & \textbf{0.77682} & 0.69671 \\ \hline
satimage-2 & 0.59438 & 0.55493 & 0.85817 & 0.97076 & 0.90414 & 0.92801 & 0.98003 & 0.96482 & 0.98159 & 0.9646 \\ \hline
vertebral & \textbf{0.48265} & 0.46904 & 0.39465 & 0.41046 & \textbf{0.48752} & 0.39953 & 0.46184 & 0.40483 & \textbf{0.50182} & 0.38861 \\ \hline
vowels & 0.77689 & 0.80236 & 0.80882 & 0.519 & 0.78853 & 0.66121 & 0.86525 & 0.52215 & 0.84647 & 0.678 \\ \hline
wbc & 0.81705 & 0.84737 & 0.67586 & 0.88776 & 0.71915 & 0.9167 & 0.81133 & 0.89594 & 0.73305 & 0.90848 \\ \hline
\end{tabular}
\caption{ROC Performance - DL methods}
\label{tab:roc-dl}
\end{table*}

\begin{table*}[!htp]
\scriptsize
\centering
\begin{tabular}{|c|c|c|c|c|c|c|c|c|c|c|c|c|c|}
\hline
Data & Samples & Dim. & Out. Perc. & ABOD & CBLOF & FB & HBOS & IForest & KNN & LOF & MCD & OCSVM & PCA \\ \hline
dl\_out & 576 & 64 & 11.1111 & \textbf{0.47533} & 0.11883 & 0.1958 & 0.10082 & 0.07764 & 0.15684 & 0.19525 & 0.47469 & 0.11109 & 0.0999 \\ \hline
2gauss\_out & 576 & 64 & 11.1111 & 0 & 0 & 0 & 0.10901 & 0.03221 & 0 & 0.004 & 0 & 0.02137 & 0.10618 \\ \hline
arrhythmia & 452 & 274 & 14.6018 & 0.38076 & 0.45385 & 0.42297 & \textbf{0.51108} & \textbf{0.49992} & 0.44637 & 0.43343 & 0.39952 & \textbf{0.4614} & 0.46129 \\ \hline
cardio & 1831 & 21 & 9.6122 & 0.23743 & 0.42966 & 0.169 & 0.44761 & 0.49186 & 0.33227 & 0.15409 & 0.42084 & 0.50112 & \textbf{0.609} \\ \hline
glass & 214 & 9 & 4.2056 & \textbf{0.17023} & 0.07262 & \textbf{0.14762} & 0 & 0.07262 & 0.07262 & 0.14762 & 0 & \textbf{0.17262} & 0.07262 \\ \hline
ionosphere & 351 & 33 & 35.8974 & \textbf{0.84415} & 0.77489 & 0.70558 & 0.32951 & 0.64743 & \textbf{0.86021} & 0.70634 & \textbf{0.88065} & 0.70005 & 0.57286 \\ \hline
letter & 1600 & 32 & 6.25 & \textbf{0.38009} & 0.23969 & \textbf{0.36419} & 0.07152 & 0.08828 & 0.33117 & \textbf{0.36411} & 0.19327 & 0.15096 & 0.08747 \\ \hline
lympho & 148 & 18 & 4.0541 & 0.44834 & 0.75167 & 0.75167 & \textbf{0.84667} & \textbf{0.87667} & \textbf{0.75167} & 0.75167 & 0.56833 & 0.75167 & 0.75167 \\ \hline
mnist & 7603 & 100 & 9.2069 & 0.3555 & \textbf{0.40231} & 0.32986 & 0.11882 & 0.30346 & \textbf{0.42043} & 0.33429 & 0.3462 & \textbf{0.39619} & 0.38461 \\ \hline
musk & 3062 & 166 & 3.1679 & 0.05075 & \textbf{1} & 0.22297 & 0.97832 & \textbf{0.98069} & 0.2733 & 0.16955 & 0.98889 & \textbf{1} & 0.97994 \\ \hline
optdigits & 5216 & 64 & 2.8758 & 0.00602 & 0 & 0.02445 & \textbf{0.2194} & 0.0271 & 0 & 0.02335 & 0 & 0 & 0 \\ \hline
pendigits & 6870 & 16 & 2.2707 & 0.08125 & 0.23974 & 0.06579 & 0.29793 & \textbf{0.35505} & 0.09844 & 0.06529 & 0.08928 & \textbf{0.32866} & \textbf{0.31865} \\ \hline
pima & 768 & 8 & 34.8958 & \textbf{0.51929} & 0.48378 & 0.44802 & \textbf{0.54238} & 0.50233 & \textbf{0.54133} & 0.45552 & 0.49625 & 0.47035 & 0.49429 \\ \hline
satellite & 6435 & 36 & 31.6395 & 0.39023 & 0.57978 & 0.39016 & 0.56903 & 0.55766 & 0.49945 & 0.38929 & \textbf{0.68452} & 0.53455 & 0.47844 \\ \hline
satimage-2 & 5803 & 36 & 1.2235 & 0.21305 & \textbf{0.93759} & 0.0555 & 0.6939 & \textbf{0.8775} & 0.38087 & 0.05551 & 0.64813 & \textbf{0.93556} & 0.80408 \\ \hline
vertebral & 240 & 6 & 12.5 & 0.06005 & 0.03381 & 0.06439 & 0.00714 & 0.05337 & 0.02381 & 0.05059 & 0 & 0.02381 & 0.02262 \\ \hline
vowels & 1456 & 12 & 3.4341 & \textbf{0.57102} & \textbf{0.36427} & 0.3224 & 0.12974 & 0.19602 & \textbf{0.50929} & 0.35506 & 0.2186 & 0.27907 & 0.13636 \\ \hline
wbc & 378 & 30 & 5.5556 & 0.30604 & 0.48064 & 0.51879 & \textbf{0.58166} & 0.50879 & 0.49518 & 0.51879 & 0.45771 & 0.51249 & 0.47673 \\ \hline
\end{tabular}
\caption{Precision @ N Performances - PyOD methods}
\label{tab:prc-pyod}
\end{table*}

\begin{table*}[!htp]
\scriptsize
\centering
\begin{tabular}{|c|c|c|c|c|c|c|c|c|c|c|c|c|}
\hline
Data & DL & SDL & \multicolumn{2}{|c|}{RKDL-S} & \multicolumn{2}{|c|}{RKDL-D} & \multicolumn{2}{|c|}{SRKDL-S} & \multicolumn{2}{|c|}{SRKDL-D} \\ \cline{4-11}
&&& rbf & poly & rbf & poly & rbf & poly & rbf & poly \\ \hline
dl\_out & \textbf{0.54116} & \textbf{0.53812} & 0.01536 & 0.02597 & 0.01429 & 0.02937 & 0.00385 & 0.01629 & 0.01087 & 0.01447 \\ \hline
2gauss\_out & \textbf{0.51047} & 0.49212 & 0.00333 & 0 & 0 & 0 & \textbf{1} & \textbf{0.78568} & 0 & 0 \\ \hline
arrhythmia & 0.42828 & 0.42783 & 0.32057 & 0.31022 & 0.35562 & 0.42643 & 0.38116 & 0.37517 & 0.37499 & 0.45478 \\ \hline
cardio & 0.30468 & 0.30867 & 0.22022 & \textbf{0.5416} & 0.19782 & 0.35779 & 0.24374 & \textbf{0.58545} & 0.1814 & 0.50126 \\ \hline
glass & 0.1369 & 0.04762 & 0.12262 & 0.09762 & 0.11429 & 0.07262 & 0.14762 & 0.03929 & 0.125 & 0.09762 \\ \hline
ionosphere & 0.8081 & 0.81393 & 0.73023 & 0.43417 & 0.78044 & 0.61622 & 0.70715 & 0.42508 & 0.72037 & 0.58191 \\ \hline
letter & 0.27955 & 0.26227 & 0.1753 & 0.01572 & 0.2086 & 0.03195 & 0.15485 & 0.02421 & 0.16185 & 0.02761 \\ \hline
lympho & 0.49833 & 0.45666 & 0.125 & 0.12833 & 0.25833 & 0.42833 & 0.22333 & 0.265 & 0.26667 & 0.465 \\ \hline
mnist & 0.37567 & 0.36151 & 0.23029 & 0.35623 & 0.13266 & 0.12771 & 0.18196 & 0.28904 & 0.16703 & 0.13247 \\ \hline
musk & 0.4075 & 0.37901 & 0.12873 & 0.23353 & 0.10472 & 0.17828 & 0.21692 & 0.68521 & 0.26918 & 0.0882 \\ \hline
optdigits & 0.00963 & 0.00696 & 0.0111 & 0 & 0.02303 & \textbf{0.04896} & 0.00474 & 0 & 0.0437 & \textbf{0.1143} \\ \hline
pendigits & 0.09973 & 0.09568 & 0.16584 & 0.28128 & 0.10907 & 0.27757 & 0.1476 & 0.27513 & 0.16232 & 0.25404 \\ \hline
pima & 0.41686 & 0.39683 & 0.43692 & 0.4744 & 0.44025 & 0.48196 & 0.45952 & 0.48992 & 0.4757 & 0.49287 \\ \hline
satellite & 0.46056 & 0.45341 & 0.47054 & 0.51423 & 0.45615 & 0.4217 & \textbf{0.61256} & 0.54732 & \textbf{0.62188} & 0.55516 \\ \hline
satimage-2 & 0.07119 & 0.04584 & 0.09132 & 0.64813 & 0.33605 & 0.38399 & 0.58768 & 0.64744 & 0.56207 & 0.62129 \\ \hline
vertebral & \textbf{0.09294} & 0.07273 & 0.06198 & 0.05927 & \textbf{0.10508} & 0.02143 & 0.05048 & 0.03214 & \textbf{0.07903} & 0.00667 \\ \hline
vowels & 0.28151 & 0.30329 & 0.22516 & 0.10201 & 0.21783 & 0.18607 & 0.25828 & 0.06196 & 0.22259 & 0.1181 \\ \hline
wbc & 0.41909 & 0.36207 & 0.18477 & \textbf{0.55319} & 0.24709 & 0.52414 & 0.27334 & \textbf{0.5301} & 0.15626 & 0.51664 \\ \hline
\end{tabular}
\caption{Precision @ N Performances - DL methods}
\label{tab:prc-dl}
\end{table*}

In general, the SDL method achieve better results than the DL method, but this is not always true. Depending on how the random selection of signals is made, there are chances that abnormal signals to be used during the training procedure. This is possible for the datasets with a very high percentage of outliers or small datasets. The same statement is valid for the KDL vs SKDL comparison. On the other hand, comparing the standard methods with the kernel methods, we notice that the second ones obtain better results. Moreover, the selective strategy improves the invariance of dictionaries to representing abnormal signals. The RKDL-D and SRKDL-D methods often improve the results. In general, the trained dictionary, $\bm{D}$, is better adapted for the representation of the normal signals. However, it is likely that the trained dictionaries contain atoms that are beneficial in the nonlinear representation of all signals, including the outliers.

The execution time of DL methods is usually larger than that of the PyOD methods. 
For example, for the {\it musk} dataset, which is among the largest, DL and SDL take about 6 seconds, i.e., not much more than MCD, which needs about 4 seconds;
RKDL algorithms take between 9 and 11 seconds, while SRKDL variant are slightly faster, with 7-10 seconds.
The other PyOD algorithms are at least $10$ times faster than the methods presented in the article.

\section{Conclusions}
In this paper we have presented a novel unsupervised method for outlier detection, based on Dictionary Learning and Kernel Dictionary Learning. We have introduced a reduced kernel DL version that is suitable for problems with large datasets. The kernel reduction technique is based on choosing a small sample of signals from the original dataset, which will further be used for the nonlinear extension. Another way to represent the kernel is to use a dictionary initially trained in with the standard DL algorithm. Both methods are accompanied by improved versions based on a random selection of the data used in the training procedure. This ensures invariance in the representation of normal signals, while the capabilities of the dictionaries for the representation of abnormal signals decrease.

Based on these results, we demonstrated that sparse learning can easily isolate the outliers from the normal signals, while obtaining competitive results with other unsupervised methods.
All the developed algorithms were introduced in an outlier detection toolbox.

\vspace*{2\baselineskip}
\bibliographystyle{unsrt}
\bibliography{bib}

\end{document}